\documentclass[lettersize,journal]{IEEEtran}
\usepackage{amsmath,amsfonts}
\usepackage{algorithmic}
\usepackage{algorithm}
\usepackage{array}
\usepackage[caption=false,font=normalsize,labelfont=sf,textfont=sf]{subfig}
\usepackage{textcomp}
\usepackage{stfloats}
\usepackage{url}
\usepackage{verbatim}
\usepackage{graphicx}
\usepackage{cite}
\usepackage{caption}
\usepackage{balance} 
\usepackage{algorithm}
\usepackage{algorithmic}
\usepackage{bm}
\usepackage{multirow}
\usepackage{graphicx}
\usepackage{booktabs} 
\usepackage[numbers]{natbib}
\usepackage[caption=false]{subfig}
\DeclareMathOperator*{\argmin}{arg\,min}

\usepackage{amsthm}

\usepackage[belowskip=0pt,aboveskip=0pt]{caption}
\begin{document}

\title{OIL-AD: An Anomaly Detection Framework for Sequential Decision Sequences}
\author{\IEEEauthorblockN{Chen Wang,
Sarah Erfani, Tansu Alpcan,
Christopher Leckie}\\
\IEEEauthorblockA{
The University of Melbourne, Melbourne, Australia\\
\{chenwang4, sarah.erfani, tansualpcan, caleckie\}@unimelb.edu.au}}



\maketitle

\begin{abstract}
Anomaly detection in decision-making sequences is a challenging problem due to the complexity of normality representation learning and the sequential nature of the task. Most existing methods based on Reinforcement Learning (RL) are difficult to implement in the real world due to unrealistic assumptions, such as having access to environment dynamics, reward signals, and online interactions with the environment. To address these limitations, we propose an unsupervised method named Offline Imitation Learning based Anomaly Detection (OIL-AD), which detects anomalies in decision-making sequences using two extracted behaviour features: \textit{action optimality} and \textit{sequential association}. Our offline learning model is an adaptation of behavioural cloning with a transformer policy network, where we modify the training process to learn a Q function and a state value function from normal trajectories. We propose that the Q function and the state value function can provide sufficient information about agents’ behavioural data, from which we derive two features for anomaly detection. The intuition behind our method is that the \textit{action optimality} feature derived from the Q function can differentiate the optimal action from others at each local state, and the \textit{sequential association} feature derived from the state value function has the potential to maintain the temporal correlations between decisions (state-action pairs).
Our experiments show that OIL-AD can achieve outstanding online anomaly detection performance with up to $34.8\%$ improvement in $F_1$ score over comparable baselines. The source code is available on https://github.com/chenwang4/OILAD
\end{abstract}


\section{Introduction}
A huge number of real-world activities can be processed as decision-making sequences, such as behaviours of autonomous vehicles \cite{sun2018probabilistic}, humans \cite{pearce2023imitating}, players in video games \cite{shao2019survey} etc. With the proliferation of RL and the availability of large-scale data, anomaly detection for decision-making sequences has become a major concern in many real-world scenarios. For example, malicious taxi drivers take detours or go through congested streets to increase their fares \cite{oh2019sequential,wang2023online}, or faulty robots perform unexpected actions resulting in safety issues \cite{haider2023out}. 

\begin{figure}[t]
\centering
\includegraphics[width=1\columnwidth]{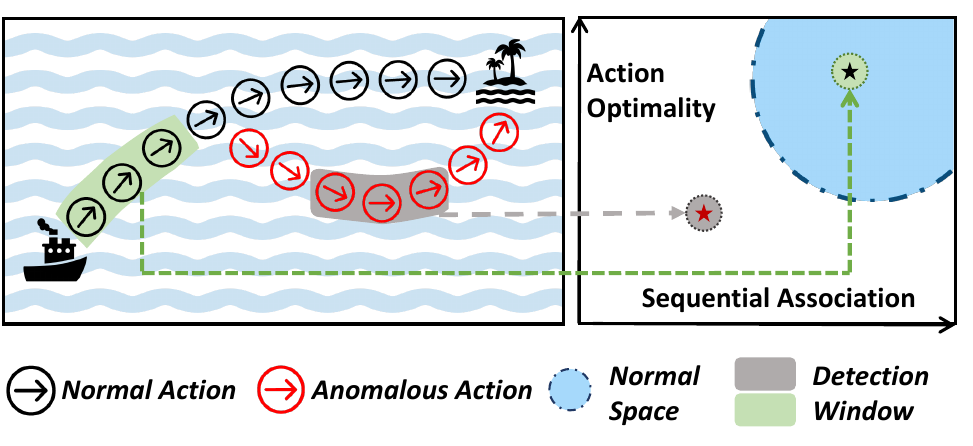} 
\caption{A demonstration of our anomaly detection method. Each decision sequence in the detection window is transformed to a novel two-dimensional feature space: action optimality and sequential association, described in Section V.D Behaviour Features for Anomaly Detection.}
\label{fig:demo}
\end{figure}

Decision-making sequences are generated by goal-oriented agents that act based on their environment and the desired goal. Intuitively, a decision-making sequence is considered anomalous if it has a different goal or unexpected events compared to other agents. Figure \ref{fig:demo} shows an example. Assume we have a set of ship trajectories and most of them follow the normal route (black arrows). The goal is to reach the island in the shortest distance. We can see that the red trajectory is an anomaly because the anomalous actions do not work towards the normal goal. This may indicate some unusual situations such as bad weather conditions.

Detecting such anomalous decision-making sequences is difficult due to the following three challenges \cite{muller2022towards}:
(1) \textit{Normality representation learning}: The idea of applying anomaly detection on a suitable data representation that can separate anomalous data from normal data has recently gained traction. Normality representation for decision-making sequences is still underdeveloped. 
(2) \textit{Sequential nature of the decisions}: Intelligent agents respond to the environment and adjust their actions accordingly to achieve an intention or a goal. In order to predict if one decision is normal or not, a good algorithm needs to take past observations and actions into consideration.
(3) \textit{Real-world implementation}: Anomaly detection methods need to be realistic and practical. In the real world, it is often infeasible to access environment dynamics, reward signals, and online interactions with the environment. 

Recent works have shown initial success in the context of RL \cite{haider2023out, zhang2021simple,sedlmeier2020policy} and Inverse Reinforcement Learning \cite{oh2019sequential, yoon2020delivering, tabatabaie2021reinforced,zhu2019sequential} albeit subject to several impractical assumptions, such as requirements for the given reward signals, knowledge of environment dynamics or access to online interactions with the environment. Therefore, these methods are difficult to apply in real-world situations. Moreover, the normality representations defined in these works (including some similarity functions between the actual state and the predicted state/state distribution, the policy entropy and the recovered reward function) are not sufficiently accurate to detect various types of anomalies in an online manner and do not consider the sequential association between new decision sequences and the history of the state-action pairs that agents have previously experienced. However, in real life, an agent's decision-making process could be influenced by its past experience. 

The goal of our work is to provide an anomaly detection method targeted at decision-making sequences, which can address all the above limitations. 
In this paper, we propose an unsupervised and online anomaly detection framework named Offline Imitation Learning based Anomaly Detection (OIL-AD), which can detect anomalies in continuous state spaces and without access to the reward signal, the environment transition probability, or any interactions with the environment (i.e., \textit{online} anomaly detection in the sense that it is based on a sliding window of observed decisions, but using \textit{offline} imitation learning in terms of not learning direct knowledge of the rewards or the environment transition probabilities). 

To learn a representation of normal decision sequences, our method learns a Q function and a state value function from normal trajectories. In RL, the Q function and the state value function build the connection between the reward function and the policy. Given the optimal policy, the Q value for the optimal action is larger than the Q values for the rest of the actions at each local state point. Moreover, from a global perspective, state values tend to be monotonically increasing if they are generated by an agent using the optimal policy \cite{wang2023online}. We define two features, action optimality and sequential association, to quantify those two properties as the normality representation, as shown in Figure \ref{fig:demo}.

To capture the sequential nature of the decision-making process, we apply a behavioural cloning 
framework with a transformer policy network. Our objective is to train the transformer policy network such that the state values computed from the state value function will be monotonically increasing for normal trajectories. By doing this, the temporal correlations between state-action pairs in decision-making are reflected in the monotonicity property. 


The main contributions of our work are as follows:
\begin{itemize}
    \item 
    We propose an offline imitation learning structure based on behavioural cloning. The proposed model can recover a Q function and a state value function and also keep the sequential nature of the decision-making process. 
    \item We define two behaviour features, action optimality and sequential association, to extract the normality representation of agents' decision-making sequences. 
    \item Our anomaly detection method is unsupervised, online and does not require access to reward signals, knowledge of the environment dynamics, or online interactions with the environment. 
    \item Our experimental results show that the anomaly detection performance of OIL-AD achieves an average improvement of around $14.15\%$ in $F_1$ score compared to existing state-of-the-art methods.
\end{itemize}

\section{Related Work}
We briefly review related works in decision sequence anomaly detection and offline imitation learning.
\subsection{Decision Sequence Anomaly Detection}
Decision-making sequences have not been a major focus of anomaly detection research. \citet{muller2022towards} formulate the anomaly detection problem in the context of RL from a conceptual perspective and propose practical desiderata for future methods, which include scarce anomalous data and the absence of a reward signal. Our method satisfies these expectations. 
\citet{zhang2021simple} design a framework based on Mahalanobis Distance to detect state outliers in an RL setting.
\citet{sedlmeier2020policy} uses the entropy of policy as the anomalous feature to detect unencountered states.
However, these methods ignore the sequential nature of the observations. \citet{haider2023out} detect out-of-distribution anomalies by comparing the actual states with the predictions from a learned dynamics model, but this method assumes that the reward signal and access to the environment are given, which could be unrealistic in practice.  

Some works incorporate Inverse RL with trajectory anomaly detection to study the decision-making sequences of drivers. Most existing methods in this line \cite{oh2019sequential, yoon2020delivering, tabatabaie2021reinforced} utilize the reward function generated by Inverse RL to infer preferences and predict behaviours of normal agents. However, these works suffer from some inherent limitations of Inverse RL, such as known environment dynamics, discrete state-action spaces, or having online interactions with the environment. Another Inverse RL-based anomaly detection method uses adversarial learning to solve the problem, with the goal of decreasing the difference between two reward functions from the generator and true anomalies \cite{zhu2019sequential}. Unlike our method, this work assumes that only anomalous data are available. 
\subsection{Offline Imitation Learning}
Offline imitation learning has attracted increasing attention in recent years due to its outstanding performance in sequential decision-making. It eliminates the requirements for access to reward signals, knowledge of the transition probabilities of the environment, and online interactions with the environment.
Therefore, it recovers the optimal policy from expert demonstrations more efficiently and realistically.

Offline imitation learning algorithms aim to match the state-action distributions of the expert and the learning agent. Energy-based distribution matching (EDM) \cite{jarrett2020strictly} formulates state distributions as an energy-based model and minimizes the KL divergence between the state-action distributions of the demonstrator and the imitator. \cite{dadashi2020primal} propose to minimize the Wasserstein distance between the expert's and the agent's state-action distributions. Inverse soft-Q learning \cite{garg2021iq} learns a single Q function that implicitly represents both reward and policy by minimizing a variety of statistical divergences between the expert and the learner. However, this distribution-matching framing considers each state-action pair as an independent sample and ignores the temporal relationships in the decision-making process.
Behavioural cloning is another method in offline imitation learning, which enjoys the benefits of simplicity and efficiency. The basic version of behavioural cloning does not involve long-term planning and suffers from compounding errors due to distribution shift \cite{ross2010efficient}. However, behavioural cloning with a Recurrent Neural Network policy network shows its ability to capture the dependencies in the decision history that agents have experienced before \cite{mandlekar2022matters}. 

Inspired by these works, we deploy a behavioural cloning framework with a transformer policy network and extract a Q function from this framework for anomaly detection. 
The difference between our work and the classical offline imitation learning methods is that our framework is designed for anomaly detection and the performance is not evaluated by average episodic rewards.  

\section{Preliminaries}
In this section, we give a brief introduction to Markov Decision Processes (MDPs) and Behavioral Cloning. 
\subsection{MDP}
MDP provides a mathematical framework for RL. MDP can be defined as a tuple $(S,A,\mathcal{P},\mathcal{R},\gamma,b_0)$, where $S$ is a state space, $A$ is an action space, $\mathcal{P}_{ss'}^a = P(S_{t+1}=s'|S_t=s,A_t=a)$ is the transition probability, $\mathcal{R}_s^a=\mathbb{E}[R_{t+1}|S_t=s,A_t=a]$ is the reward, $\gamma \in [0,1)$ is the discount factor and $b_0(s)$ is the probability of starting at state $s$. At each state $s$, the agent selects an available action $a$, and the environment will take the agent to the next state $s'$ according to the transition probability $\mathcal{P}$. Meanwhile, the agent will be given a corresponding reward $\mathcal{R}_s^a$.
The goal of MDP is to find an optimal policy $\pi^*$ to maximize the expected value of discounted future rewards. A policy $\pi$ is a distribution over actions given a state, expressed as $\pi (a|s) = P[A_t = a|S_t = s]$. 
Each policy is related to one state-value function $v_\pi (s)$ and one action-value function $Q_\pi (s,a)$. According to the Bellman expectation equation, we can have:
$v_\pi (s)=\mathbb{E}_{\pi}[\mathcal{R}_{t+1}+\gamma v_\pi(S_{t+1})|S_t=s]$ 
and $Q_\pi (s,a) = \mathbb{E}_{\pi}[\mathcal{R}_{t+1}+\gamma Q_\pi(S_{t+1},A_{t+1})|S_t=s, A_t=a]$.
An optimal policy $\pi^*$ can be found by maximizing over the action-value function $Q_* (s,a)=\max_{\pi} Q_{\pi}(s,a)$. 

\subsection{Imitation Learning and Behavioural Cloning}
In the setting of Imitation Learning, the reward is not given. Instead, the goal is to learn a behaviour policy from a set of expert demonstrations $D:=\{(s_k,a_k)\}_{k=1}^K$ (where $K$ is the number of demonstrated state-action pairs). This setting is more realistic for some applications where it is impractical to pre-specify an accurate reward function. 

Behavioural cloning is one of the simplest methods of Imitation Learning and is widely used in practice \cite{li2022rethinking}.
Unlike Inverse RL, behavioural cloning can learn the optimal policy without the intermediate step of learning a reward function. The standard behavioural cloning algorithm learns a mapping of states to actions from expert demonstrations using supervised learning to minimize the difference between the observed expert actions and the predicted actions:
\begin{equation}
    \argmin_\theta \mathbb{E}_{(s,a)\sim D} L(a, \pi_\theta (s))
\end{equation}
where $L$ is the cost function and $\theta$ represents the policy network parameters.

Behavioural cloning has massive advantages in terms of simplicity and efficiency, but it also suffers from the difficulty of recovering from distribution shifts \cite{zhan2018generative, hua2021learning}. However, the robustness to distribution shifts is not part of the performance evaluation when we switch our perspective from offline imitation learning to the problem of anomaly detection. Furthermore, the compounding error can even potentially help to detect abnormal patterns (out-of-distribution) from training/normal data.

\section{Problem Statement}
We consider RL agents in an MDP environment, defined by a tuple $(S, A, \mathcal{P}, \mathcal{R}, \gamma, b_0)$. Goal-oriented agents take actions to maximize their sum of future rewards $\sum_{t}^{t\to\infty}\gamma^tr(s_t,a_t)$. Each decision-making sequence $\tau$ with a variable length $T$ is composed of state-action pairs $\{(s_1,a_1),\dots, (s_T,a_T) \}$. In our anomaly detection setting, let $C_{normal}$ and $C_{anomaly}$ denote the sequence sets for normal and anomalous agents respectively.
\begin{equation}
    C_{normal}\cap C_{anomaly} = \emptyset \quad \land \quad C_{anomaly} = C \setminus C_{normal}
\end{equation}
We consider that anomalous behaviours are performed because of anomalous intention $\mathcal{R}_{anomaly}$ and unexpected transition $\mathcal{P}_{anomaly}$. In this work, we do not know any information from the environment such as $\mathcal{R}$ , $\mathcal{P}$, or $\gamma$, and we can only observe decision-making sequences of agents from $C_{normal}$.
The goal is to extract an appropriate feature representation $\phi(\tau)$ from sequences, and find a suitable detection model to predict the anomalous level $p$ such that
\begin{equation}
    p(\phi(\tau)) < p(\phi(\tau')) \quad \forall \tau \in C_{normal}, \space \forall \tau' \in C_{anomaly}
\end{equation}
\section{Methodology}
In this section, we propose an anomaly detection framework for sequential decision sequences. We start with two desired properties for our model to recover the Q function and the state value function. Then we provide the training process for the proposed model. Lastly, we define two behavioural features to distinguish anomalies from normal trajectories.
\begin{figure*}[t]
\centering
\includegraphics[width=2\columnwidth]{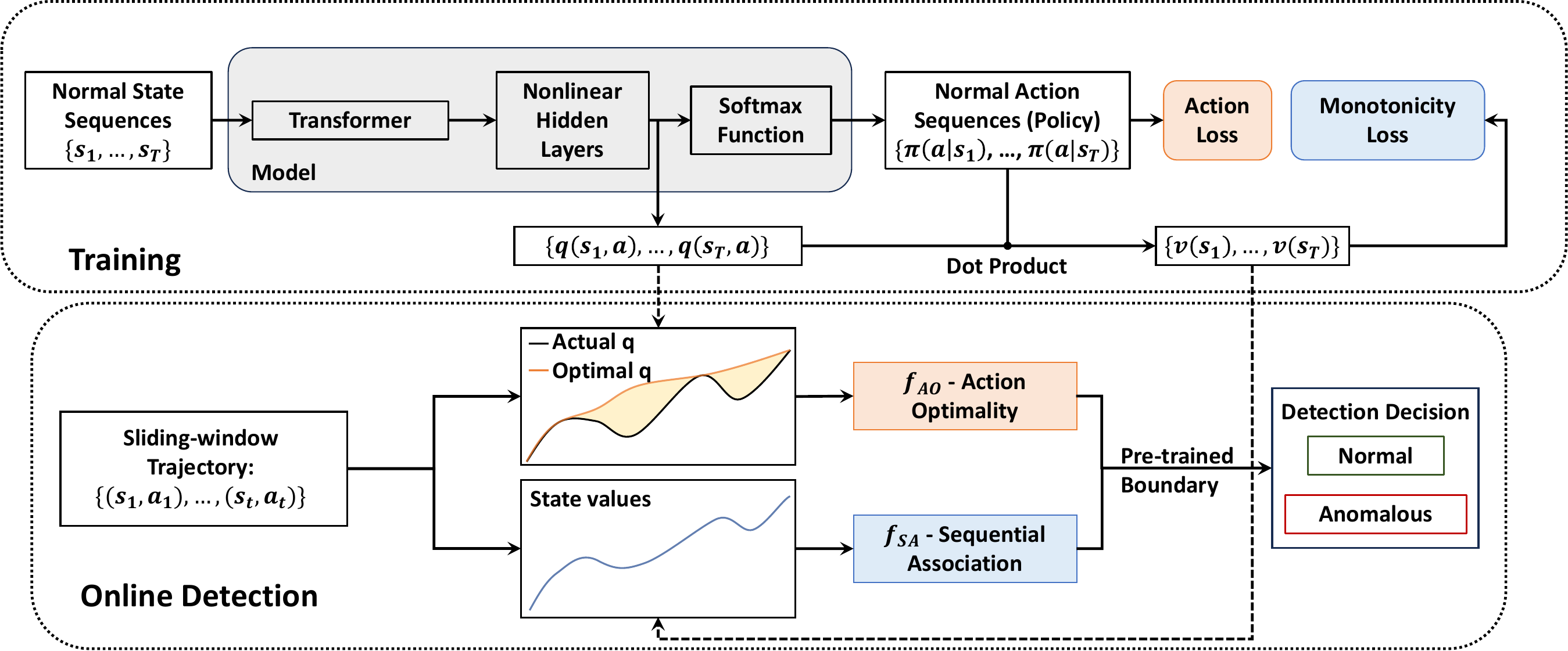} 
\caption{Method overview. In the training stage, the model is updated based on the action loss and monotonicity loss. These two training objectives contribute to the extracted features for detection - action optimality and sequential association, respectively.}
\label{fig2}
\end{figure*}
\subsection{Intuitions}
Our goal is to learn a Q function and a state value function that are close to the ground truth. Therefore, we propose two properties: (1) the Q function can select the optimal/normal action in a given state, and (2) the state values for optimal/normal trajectories tend to be monotonically increasing. 

The Q function represents the expected total reward of taking a specific action in a given state, which provides a quantitative measure for the values of different actions. By estimating the Q-values of all possible actions in each state, we expect that an agent can determine which actions are most likely to lead to high rewards and use this information to select actions that optimize its long-term performance.
\begin{equation}
    q(s,a^*) > q(s,a), \quad \forall a \neq a^*, \forall s \in S
    \label{equ:req1}
\end{equation}
where $a^*$ is the optimal action.

In principle, the state value function satisfies the Bellman equation, that is, a recursive relationship between the values in one period and the values in the next period. This property allows those two functions to be updated iteratively during RL, using the rewards obtained from the environment to refine its estimates of the expected values of different actions. 
However, in the context of offline imitation learning, it is impossible to recover the true state value function because it is not feasible to guarantee the Bellman equation without knowing the transition probability or having access to the environment. \citet{wang2023online} proposed the monotonicity property of state values based on the Bellman equation: that state values of trajectories performed by the optimal policy should tend to be monotonically increasing.
Therefore, we use that as our second expected property. 
\begin{equation}
    v_\pi(s) = \sum_a \pi(a|s)q_\pi(s,a)
    \label{equ:vandq} 
\end{equation}
\begin{equation}
    v_\pi(s_1) \leq v_\pi(s_2) \leq \dots \leq v_\pi(s_T), \quad \{s_i\}_{i=1}^{T} \in \tau^*
    \label{equ:req2}
\end{equation}
where $\tau^*$ is a sequence that follows the optimal policy.
This modified property maintains the connection between the state value function $v(s)$ and the Q function, and it is derived from the Bellman equation, which can be seen as an approximation to satisfy the Bellman equation. 
\subsection{Monotonicity of State Values}
\citet{wang2023online} proposed the monotonically increasing property of state values. We believe this property can be beneficial to approximate the true state value function.
Here, we provide an extended proof of \cite{wang2023online} in the offline imitation learning setting and our remark of a special case.

\textbf{Theorem 1.}
\textit{Assume a stationary environment with no random events or noise. Let $s_t$ be a state in a normal trajectory at time $t$. Let $T$ be the length of the trajectory. Let $\pi$ be the optimal policy that performs the specific normal trajectory. For any MDP that satisfies $\mathbb{E}_{\pi}[r_{t+1}] \leq \mathbb{E}_{\pi}[\sum_{k=t+2}^T (1-\gamma)\gamma^{k-t-2}r_k ]$, the corresponding value function $v_{\pi}$ has $v_{\pi}(s_t) \leq v_{\pi}(s_{t+1}) \forall t \in [0,T)$.} 
\begin{proof}
By definition in RL \cite{sutton2018reinforcement}, we have
\begin{equation}
      v_{\pi}(s) = \mathbb{E}_{\pi}[G_t|S_t = s]   
\end{equation}
where $G_t$ is the expected discounted return.
Then, $\forall s_t \in S$
\begin{equation}
    v_{\pi}(s_t) 
    =\mathbb{E}_{\pi}[\sum_{k=t+1}^T \gamma^{k-t-1}r_k|S_t=s_t]
\end{equation}
Therefore,
\begin{equation}
\begin{aligned}
     &v_{\pi}(s_t) -  v_{\pi}(s_{t+1})\\
     &= \mathbb{E}_{\pi}[\sum_{k=t+1}^T \gamma^{k-t-1}r_k] - \mathbb{E}_{\pi}[\sum_{k=t+2}^T \gamma^{k-t-2}r_k] \\
     &=  \mathbb{E}_{\pi}[r_{t+1} + \sum_{k=t+2}^T \gamma^{k-t-1}r_k - \sum_{k=t+2}^T \gamma^{k-t-2}r_k] \\
     &=  \mathbb{E}_{\pi}[r_{t+1} + \sum_{k=t+2}^T (\gamma -1)\gamma^{k-t-2}r_k ]
\end{aligned}
\end{equation}
where $\gamma \in [0,1)$.
Then we can derive the condition for MDPs in which $\pi$ is the optimal policy and $v_{\pi}(s_t) \leq  v_{\pi}(s_{t+1})$ :
\begin{equation}
    \label{equ:condition}
    \mathbb{E}_{\pi}[r_{t+1}] \leq \mathbb{E}_{\pi}[\sum_{k=t+2}^T (1-\gamma)\gamma^{k-t-2}r_k ]
\end{equation} 
\end{proof}
\textit{Remark:}
The condition in Eq. \ref{equ:condition} can be further simplified in a setting where the goal is to reach the destination as fast as possible, such as in Mountain Car environment \cite{1606.01540}, Acrobot Environment \cite{1606.01540}, Gridworld environment \cite{sutton2018reinforcement} and etc.
In this case, the reward signal is a negative constant for each timestep till the agent achieves the goal. 
\begin{proof}
    Given a deterministic environment, for each trajectory following the optimal policy, Eq. \ref{equ:condition} becomes:
\begin{equation}
    r_{t+1} \leq \sum_{k=t+2}^T (1-\gamma)\gamma^{k-t-2}r_k 
    \label{equ:condition_determ}
\end{equation}
If $r_{t}=r<0, \quad \forall t\in[0,T]$, then we have
\begin{equation}
    r \leq r(1-\gamma^{T-t-2})
\end{equation}
\begin{equation}
\label{equ:nagative_r}
    1 \geq 1-\gamma^{T-t-2}
\end{equation}
If $r_{t}=r<0, \quad \forall t\in[0,T)$ and $r_T>0$ (the final reward for achieving the goal is a positive value), then Eq. \ref{equ:condition_determ} becomes: 
\begin{equation}
\label{equ:rT>0}
    r \leq r(1-\gamma^{T-t-3}) + (1-\gamma)\gamma^{T-t-2}r_T
\end{equation}
Both Eq. \ref{equ:nagative_r} and \ref{equ:rT>0} are always true. Therefore, state values of optimal trajectories are theoretically guaranteed to be monotonically increasing in an MDP environment where the reward signal is a negative constant for each timestep till the agent achieves the goal.
\end{proof}

\subsection{Model Structure and Training Process}
The model structure is shown in Figure \ref{fig2}. We use a variant of traditional behavioural cloning that adds a transformer block in the policy network to maintain the sequential nature of decision-making. This model can be used in continuous state space and discrete action space. 
\begin{algorithm}[tb]
\caption{Model training}
\label{alg:training}
\textbf{Input}: Normal trajectories $C_{normal}$, regularisation parameter $\alpha$, training iterations for action loss $n$ and training iterations for monotonicity loss $n_1$\\
\begin{algorithmic}[1] 
\STATE Get normal state and action sequences from $C_{normal}$
\STATE Initialize policy network $\pi_\theta$
\FOR{iterations i = 1 to n}
\STATE Update parameters $\theta$ using Eq.\ref{equ:first obj} \hfill\COMMENT{ Action Loss}
\IF {i $>n_1$}
\STATE Update parameters $\theta$ using Eq.\ref{equ:second obj} \hfill\COMMENT{ Monotonicity Loss}
\ENDIF
\ENDFOR
\STATE \textbf{return} policy network $\pi_\theta$, Q function $q(s,a,\theta)$ and state value function $v(s,\theta)$
\end{algorithmic}
\end{algorithm}
To construct an expected Q function, we consider the values before the softmax layer as Q values $\{q(s,a)\}_{a\in A}$ for one state $s$.
In this way, the policy is an exponential soft-max distribution:
\begin{equation}
    \pi(a|s,\bm{\theta}) = \frac{e^{q(s,a,\bm{\theta})}}{\sum_b e^{q(s,b,\bm{\theta})} }
\end{equation}
where $\bm{\theta}$ is the policy network parameters. By computing the dot product between $q(s,a)$ and $\pi(a|s)$ as Eq. \ref{equ:vandq}, we can derive the state values $v(s)$. 

We use two training objectives to make the model satisfy the two properties in Eq.\ref{equ:req1} and Eq.\ref{equ:req2} respectively. The first training objective, named action loss, is formulated as:
\begin{equation}
    \min_{\bm{\theta}} H(\pi^*(a|s)|\pi(a|s,\bm{\theta})) - \alpha H(\pi(s,a,\bm{\theta}))
    \label{equ:first obj}
\end{equation}
where $H$ represents the entropy function. The first term is the cross entropy of the predicted actions $\pi(a|s)$ and true actions $\pi^*(a|s)$. The second term is the self-entropy of the predicted actions $\pi(a|s)$, and $\alpha>0$ here is a hyper-parameter to decide the level of regularisation. We add the second term to prevent overfitting. 
After training on the normal agent trajectories, we can obtain a policy neural network and achieve the first property for the Q function.

In this work, we extract the monotonic information by using Spearman's coefficient. More specifically, for each trajectory $\tau\in C_{n}$, we can derive state values $\bm{v}_\tau=\{v_{\pi_{\bm{\theta}}}(s_1),v_{\pi_{\bm{\theta}}}(s_2),\dots,v_{\pi_{\bm{\theta}}}(s_T)\}$ and time steps $\bm{t}_\tau=\{1,2,\dots,T\}$. The second objective, named monotonicity loss, is to maximize Spearman's rank correlation coefficient between $\bm{v}_\tau$ and $\bm{t}_\tau$:
\begin{equation}
    \max_{\bm{\theta}} \frac{cov(rank(\bm{v}_\tau),rank(\bm{t}_\tau))}{\sigma_{R(\bm{v}_\tau)}\sigma_{R(\bm{t}_\tau)}}
    \label{equ:second obj}
\end{equation}
where $rank(\bm{v}_\tau)$ and $rank(\bm{t}_\tau)$ are the ranks of $\bm{v}_\tau$ and $\bm{t}_\tau$ respectively, $cov$ is the covariance of the ranks, and $\sigma_{R(\bm{v}_\tau)}$ and $\sigma_{R(\bm{t}_\tau)}$ are the standard deviations of the ranks.
The prior knowledge of monotonicity can enhance the performance of models and many researchers are working on incorporating monotonicity in deep neural networks. Recent works try to change the model architecture \cite{yanagisawa2022hierarchical} or the training process \cite{gupta2019incorporate,monteiro2022monotonicity} to introduce the monotonicity between a subset of input features and the output. However, we are not seeking the monotonicity between the input features and the output. In other words, we cannot use the time steps as one input feature and build the simple monotonicity property between time and state values. Instead, the model needs to learn the monotonicity from the temporal correlation in the state-action sequences. The reason is that time steps are invariant for both normal and anomalous trajectories and we expect the state values only for normal trajectories to be monotonically increasing and the state values for anomalies do not follow this trend. Spearman's coefficient is widely used to evaluate the monotonicity property \cite{zhou2020bp,zhan2023long}. Since a fast differentiable Spearman's rank correlation coefficient was proposed by \citet{blondel2020fast}, it has become feasible to utilize it as a loss function for deep neural networks. 

The training process is shown in Algorithm \ref{alg:training}. We first train the policy neural network based on Eq.\ref{equ:first obj} to learn the optimal policy as a good foundation for the next stage. This step is similar to the regular training in Behavioural Cloning.
To recover the temporal relationship, we then train the model based on Eq.\ref{equ:first obj} and Eq.\ref{equ:second obj} at the same time. Based on our experiments, we find that half of the training iterations $\frac{n}{2}$ is generally a good choice for $n_1$ (the number of iterations for the second stage).

\subsection{Behaviour Features for Anomaly Detection}
At the detection stage, we have access to the learned Q function and state value function from the trained model.
We define two features to extract the behavioural pattern from the trajectories.
\begin{enumerate}
    \item \textbf{Action optimality} ($f_{AO}\leq0$): the negative area between actual $q$ values and predicted optimal $q$ values. This feature characterises whether the agent is selecting the normal/optimal action at each state. We apply the algorithm from \cite{jekel2019similarity} to calculate the area between the two curves. The computed area can not only capture the anomalous actions the agents make, but also how anomalous these actions are. For example, at some states, the agent could have one optimal action, one sub-optimal action and a few poor actions: $q(s,a^*)>q(s,a_{sub})\gg q(s,a_{poor})$. The sub-optimal action $a_{sub}$ is less anomalous than these poor actions $a_{poor}$. 
    \item \textbf{Sequential association} ($f_{SA}\in[-1,1]$): a Spearman's correlation coefficient between state values and the time step sequence to show if state values are monotonically increasing. This feature characterises whether the agent is making the right sequential decisions given the context of desired normal goals.
\end{enumerate}
In the ideal case, we have $f_{AO}=0$ and $f_{SA}=1$ for a normal trajectory.
\subsection{Online Detection}
We use the sliding window method to achieve online detection. We define $w_q$ and $w_v$ as the window lengths for action optimality feature $f_{AO}$ and sequential association feature $f_{SA}$ respectively. If $w_q$ and $w_v$ are not equal, the features with the smaller window length will be downsampled and the maximal value will be chosen to represent itself in the larger window.  

Each windowed trajectory can be represented by two features $(f_{AO},f_{SA})$ in the latent space. We apply the Isolation Forest method \cite{liu2008isolation} to create a boundary in the two-dimensional latent space using the features generated from the training dataset. The choice of the contamination parameter of the Isolation Forest is very close to zero in this case. 
Note that in this process, we do not need to have access to the real anomalous trajectories, which maintains the property of unsupervised learning and one-class classification. 

\subsection{Algorithm Overview}
Given a normal trajectory dataset $\Gamma_n$, we can separate state sequences and the corresponding action sequences. We use state sequences as the input and action sequences as the expected outputs/labels, and train a behavioural cloning model with a transformer policy network in a supervised way to recover the optimal policy $\pi_\theta$. By extracting the values before the softmax layer and applying Spearman's coefficient as the monotonicity loss function, we also learn the Q function $q(s,a,\theta)$ and state value function $v(s,\theta)$ for normal agents. After the offline training, we can extract the action optimality feature and sequential association feature from each windowed trajectory. We first generate the boundary from the normal features and then we can transform any new windowed trajectory to the latent space $(f_{AO},f_{SA})$. If the behaviour features in the latent space exceed the pre-trained boundary, the trajectory is identified as an anomaly; or otherwise, as normal.

Our proposed method aims to solve anomaly detection problems targeted at decision-making sequences in a more realistic setting (with no access to reward signals, no knowledge of environment dynamics, and no online interactions with the environment). We define two features to represent the behavioural normality based on the learned Q function and the state value function. The intuition is that the learned Q function can distinguish the normal actions following the optimal policy from others and the state values for each normal trajectory tend to monotonically increase with time.

\begin{figure*}[t]
    \centering
  \subfloat[Policy Anomaly\label{Policy Anomaly}]{%
       \includegraphics[width=0.45\linewidth]{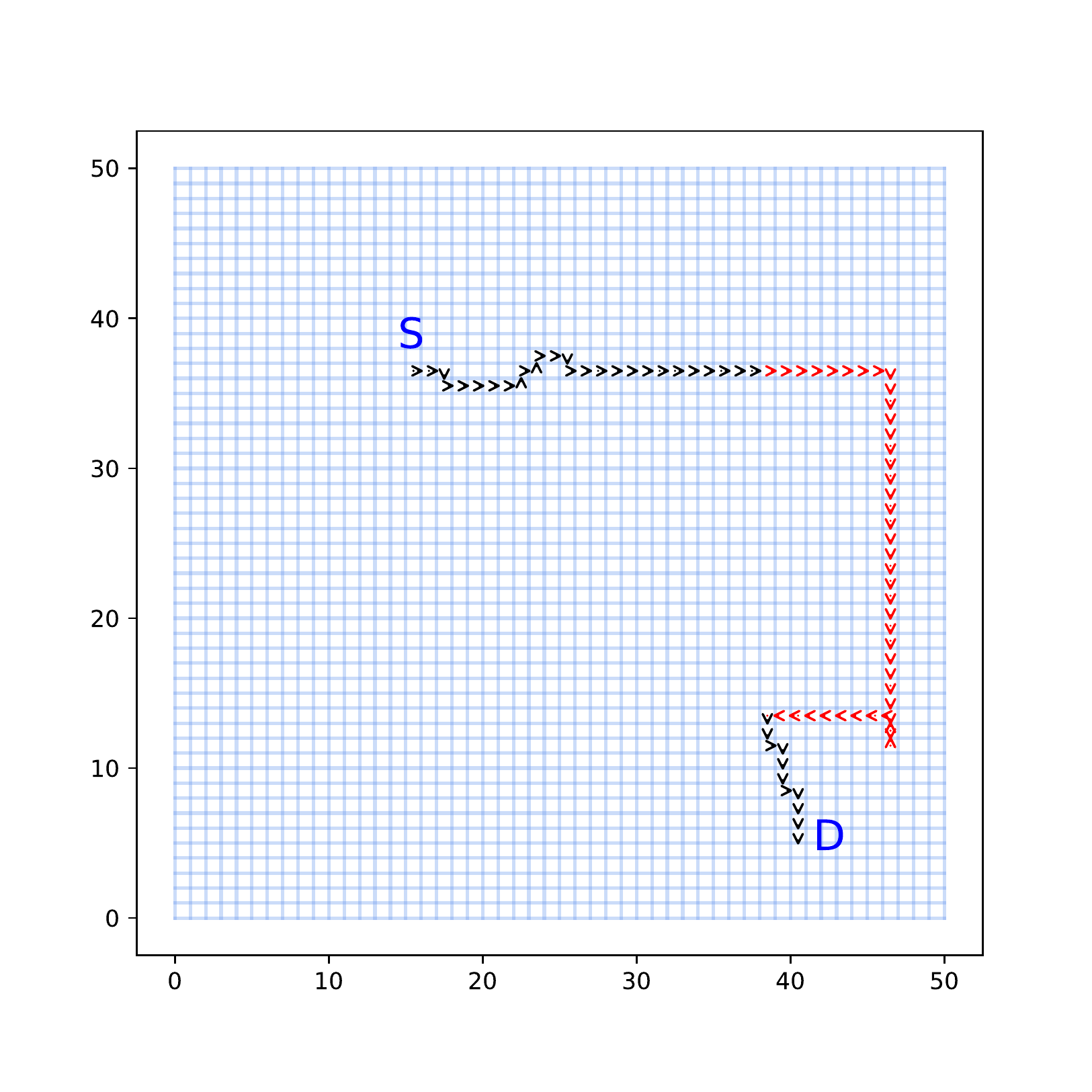}}
    \hspace{0em}
  \subfloat[Perturbed Anomaly\label{Perturbed Anomaly}]{%
        \includegraphics[width=0.45\linewidth]{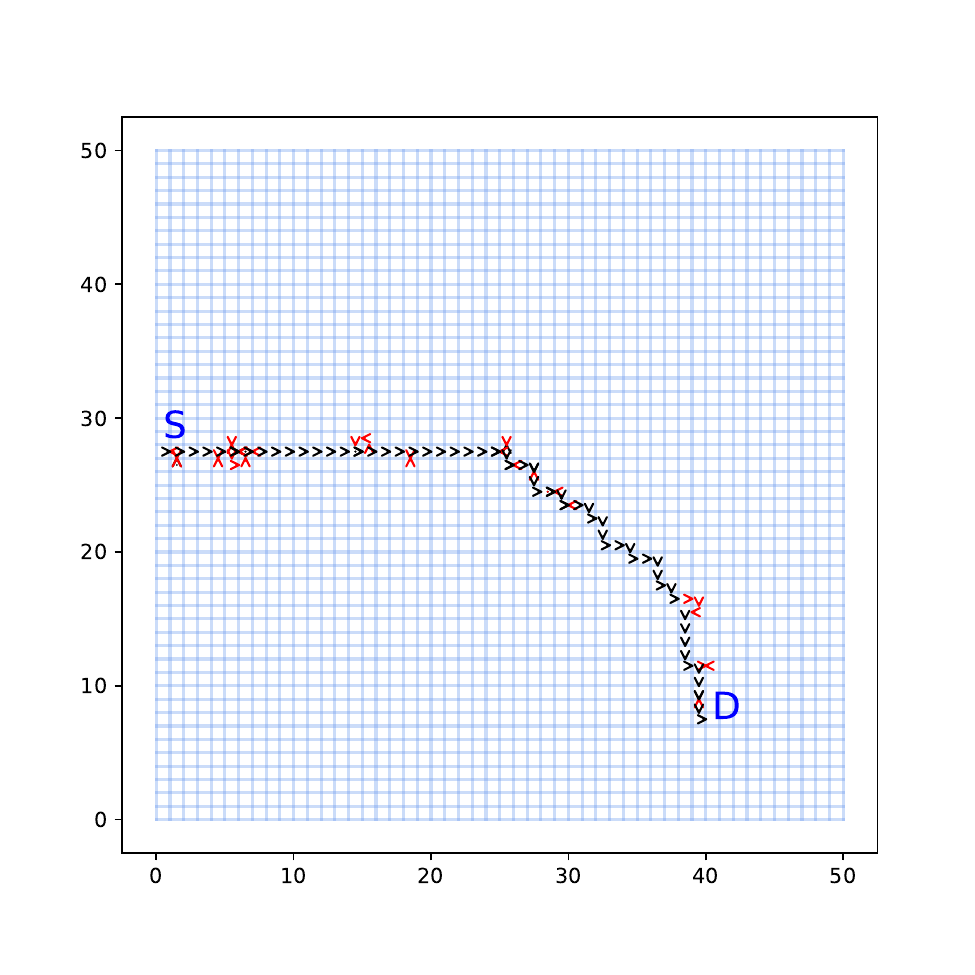}}

  \caption{Examples of generated anomalous trajectories from the Chengdu dataset in a $50*50$ grid map. The black arrows indicate normal behaviour and the red arrows indicate anomalous behaviour. $S$ and $D$ represent source and destination respectively.}
  \label{fig:trajs}
\end{figure*}
\begin{figure*}[t]
    \centering
  \subfloat[State Values\label{State Values}]{%
       \includegraphics[width=0.25\linewidth]{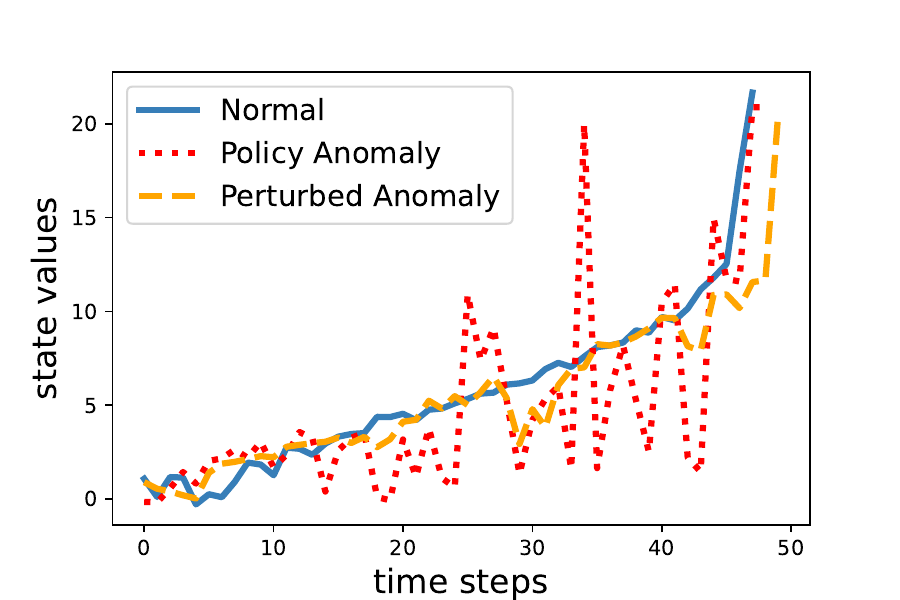}}
    \hfill
    \subfloat[Q-Normal\label{Normal}]{%
        \includegraphics[width=0.25\linewidth]{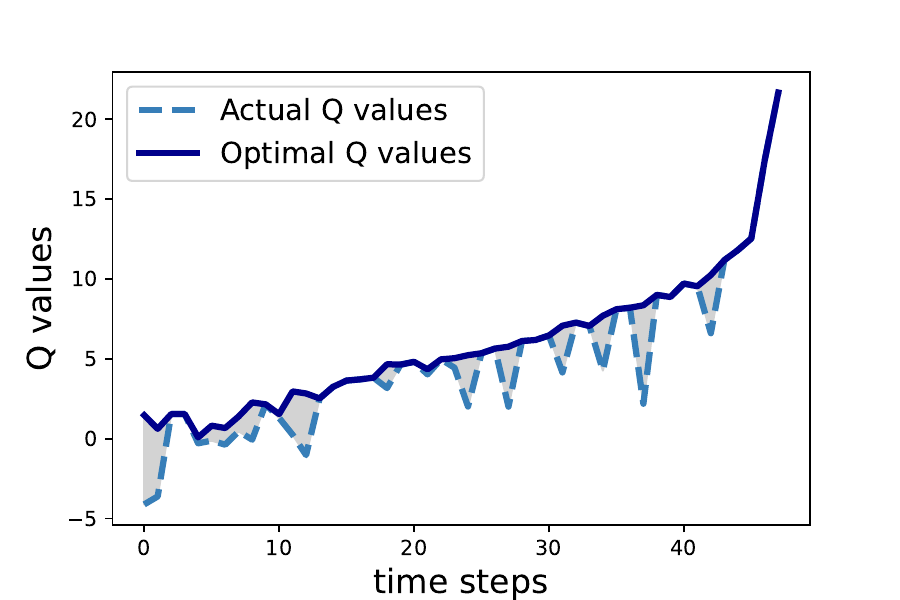}}
    \hfill
    \subfloat[Q-Policy Anomaly\label{Detour}]{%
        \includegraphics[width=0.25\linewidth]{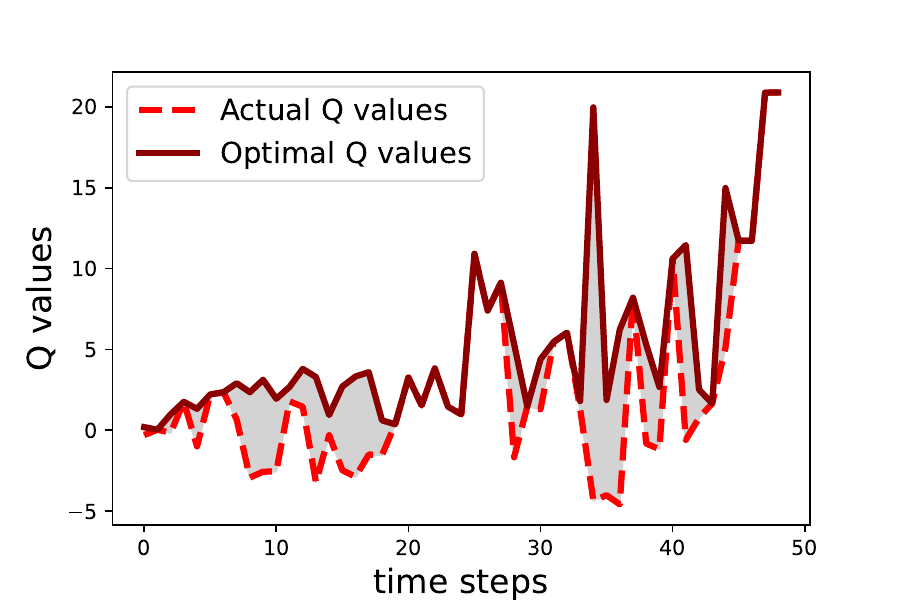}}
        \hfill
  \subfloat[Q-Perturbed Anomaly\label{Perturbed }]{%
        \includegraphics[width=0.25\linewidth]{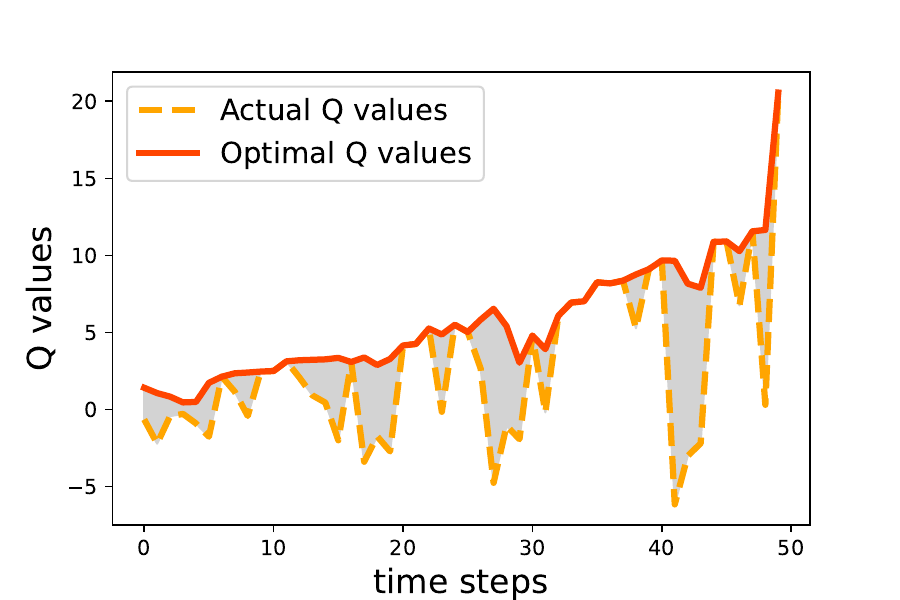}} 
  \caption{Generated state values and Q values of one normal trajectory, one policy anomalous trajectory and one perturbed anomalous trajectory. Examples are from the Chengdu dataset.}
  \label{fig:qv_plots_examples}
\end{figure*}
\section{Experiments}
In our experiments, we focus on three research questions: (1) How does OIL-AD compare with the baselines in terms of different types of anomalies? (2) How effective are our proposed behaviour features in generating the representation of sequential decision sequences? and (3) How important is each individual objective function of OIL-AD? (4) Is the recovered state value function close to the ground truth?

\subsection{Experimental Setup}
We apply our method and other baselines to three different datasets including two real-life datasets and one simulated dataset from the Gym environment \cite{1606.01540}. To provide a comprehensive evaluation, we generate two types of anomalies, policy anomalies and perturbed anomalies, for each dataset.
\subsubsection{Datasets}
\textbf{Real-life Datasets:}
1) \textit{Chengdu Dataset} \cite{didi} contains taxi trajectories within the Second Ring Road of Chengdu City in China. The time interval between GPS data points is approximately 2-4 seconds. The record for each point has five features: driver ID, order ID, timestamp, longitude, and latitude. During the pre-processing step, we extract trajectories with the same order ID. We transform GPS data into state-action pairs in a discrete grid world ($50*50$) based on MDP. Therefore, this dataset has a discrete state space and discrete action space.

2) \textit{AIS Dataset} \cite{AIS} contains sea vessel traffic data from sub-areas in Australia. Each point recorded in these datasets is a vessel position report. Each record point includes craft ID, longitude, latitude, course, speed, vessel type, vessel sub-type, vessel length, beam, draught and timestamp. During the pre-processing step, we extract trajectories that are from cargo and tanker vessels from July to December 2020 in the Bass Strait area. Each state includes latitude, longitude and speed. We discretize the course into five actions: moving north, moving east, moving south, moving west and staying stationary. This dataset has a continuous state space and discrete action space. 

\textbf{Gym Dataset:}
3) \textit{LunarLander} \cite{1606.01540} contains trajectories generated by a well-trained Proximal Policy Optimization (PPO) \cite{schulman2017proximal} agent in the gym environment \textit{LunarLander-v2} that solves a classic rocket optimization problem. The average expected reward of each trajectory from the demonstrator is $281.18\pm22.93$.
This dataset has a continuous state space and discrete action space. 

In terms of the difficulty of the datasets, intuitively, we believe the AIS dataset is more challenging than the Chengdu dataset, and both are more challenging than the LunarLander dataset. The LunarLander dataset is synthetic and has one single intention. The Changdu dataset is multi-intention but constrained by roads. The AIS dataset is multi-intention, less constrained compared with the Chengdu dataset, and its state space is continuous. 
\subsubsection{Anomaly Generation}
1) \textit{Policy Anomalies} are generated by agents that have a different policy or intention ($R$) than the normal agents. For the Chengdu and AIS datasets, we generated detour deviations based on normal trajectories, as shown in Figure \ref{Policy Anomaly}. We set two parameters to control the deviation: deviation $d$ and the proportion $\alpha$. More specifically, a proportion $\alpha$ of a normal trajectory is replaced with a detour deviation that has extra distance $d$ compared with the original normal segment. For the Chengdu dataset, we set $d=20$ grid steps and $\alpha=0.6$. For the AIS dataset, we set $d=2$ degrees and $\alpha=0.6$.
For the LunarLander dataset, the policy anomalies are generated by sub-optimal agents where the anomalous level is quantified by the average expected rewards of each trajectory, where lower rewards indicate a more anomalous trajectory. The sub-optimal agents used in LunarLander have an anomalous level of $56.94\pm90.38$. 

2) \textit{Perturbed Anomalies} are generated by adding small random perturbations on normal trajectories, which means the normal transition probability is disrupted. For the Chengdu dataset with a discrete state space, the perturbed anomalies share the same policy with the normal agents but with a small probability $\omega$ to choose a random action instead of the optimal action, as shown in Figure \ref{Perturbed Anomaly}. The random probability $\omega$ is set to 0.4.
For datasets with continuous state spaces (AIS and gym datasets), we add Gaussian noise to the state measurements of normal trajectories. We set the Gaussian noise to $N(0, 0.05)$ for the AIS dataset, and $N(0, 0.1)$ for the LunarLander dataset. 

\subsubsection{Baselines}
We compare our model with an extensive variety of baselines, including classic anomaly detection methods: One-Class Support Vector Machine (OC SVM), Isolation Forest (IF), Local Outlier Factor (LOF); state-of-the-art probabilistic model: Empirical-Cumulative-distribution-based Outlier Detection (ECOD) \cite{li2022ecod}; and deep learning models: Long Short Term Memory Autoencoder (LSTM-AE) \cite{malhotra2016lstm}, Gaussian Mixture Variational Sequence AutoEncoder (GM-VSAE) \cite{liu2020online}, TS2vec \cite{yue2022ts2vec} (without adjusted anomaly results). 

\subsubsection{Evaluation Metrics}
For each experiment, we record the number of true positives (TP), false positives (FP), true negatives (TN), and false negatives (FN). TP refers to anomalous trajectories that are also detected as anomalies, FP refers to normal trajectories that are detected as anomalies, TN refers to normal trajectories that are also detected as normal and FN refers to anomalous trajectories that are detected as normal. To compare anomaly detection performance for each anomaly type, we use precision, recall, and $F_1$ score as evaluation metrics: $\textrm{precision} = \frac{\text{TP}}{\text{TP}+\text{FP}}$, $\textrm{recall} = \frac{\text{TP}}{\text{TP}+\text{FN}}$ and $\text{F}_1 = \frac{2*\text{precision}*\text{recall}}{\text{precision}+\text{recall}}$.
The parameters for all models are tuned to achieve the best $F_1$ score performance.

\subsection{Implementation Details}
Models are trained using the AdamW optimizer \cite{loshchilov2018decoupled}. The learning rates for the action loss function are $10^{-3}$, $10^{-3}$ to $10^{-7}$ following Pytorch cyclical learning rate scheduler, $5*10^{-3}$ for Chengdu, AIS, and LunarLander datasets respectively. The learning rates for the monotonicity loss function are $2*10^{-4}$, $10^{-4}$, and $10^{-4}$ for Chengdu, AIS, and LunarLander datasets respectively.
All training is done within 12 epochs. Other hyperparameters are shown in Table \ref{tab:hyper}. 
At the testing stage, each testing trajectory was randomly selected from a large data pool and the process was repeated 5 times to derive the mean values. All experiments were performed on a 3.2GHz Apple M1 with 16 GB RAM running Mac OS X 13.
\begin{table}[htp]
\begin{center}
	\caption{Hyperparameters of OIL-AD}
	\label{tab:hyper}
	\begin{tabular}{l|c|c|c}\toprule
		Hyperparameter & Chengdu & AIS &LunarLander\\ \midrule
	    Transformer layers&1&3&1\\
            Emedding dimension &128&128&256\\
            Attention heads&2&1&4\\
            Dropout&0.1&0.1&0.1\\
            Sliding window size $w_q$&20&40&20 \\
            Sliding window size $w_v$ &15&40&60\\
            Step size& 1&1&1\\
            Regularisation parameter $\alpha$& 0.05& 0 &0.1 \\
            Contamination (IF)&  0.008&0.045 & 0.00035\\
            Training data size& 6023& 1291&8000\\
            Anomaly rate& 14.3$\%$& $9.1\%$ & $5\%$\\
		 \bottomrule
	\end{tabular}
\end{center}
\end{table}

\subsection{Evaluation on Baselines}

As shown in Table \ref{tab:main results}, our method OIL-AD consistently achieves the best performance in terms of $F_1$ measure, by a significant margin on all datasets. Figure \ref{fig:qv_plots_examples} shows the effectiveness of our anomaly detection method. State values of the normal trajectory tend to be monotonically increasing (Figure \ref{State Values}), and the gap between the actual Q values and optimal Q values from the normal trajectory (Figure \ref{Normal}) is less than the anomalies (Figure \ref{Detour} and Figure \ref{Perturbed }).

In the Chengdu dataset, normal trajectories are limited to three different destinations. Our method OIL-AD achieves improvements in $F_1$ score up to $15.2\%$ and $34.8\%$ for policy anomalies and perturbed anomalies respectively.
Deep learning baselines, such as GM-VASE and LSTM-AE, achieve the second and the third highest $F_1$ score for policy anomalies, while they have relatively poor performance for detecting perturbed anomalies. This indicates that these deep learning models are able to learn the sequential relationship but they may not be robust to small perturbations. 

In the AIS dataset, the vessel behaviours have wide variance and different intentions. For example, the trajectories can have different and unlimited numbers of destinations. The performance from all the models drops in this dataset. However, OIL-AD still outperforms all the baselines by at least $8.3\%$ and $21.8\%$ in $F_1$ score on policy anomalies and perturbed anomalies respectively, suggesting that our model has the potential to capture the sequential behavioural characteristics in a challenging real-life environment that includes continuous state space, multiple intentions and noise. 

Unlike real-life datasets, the normal trajectories in the LunarLander dataset do not have noise and they come from one intention/reward function. OIL-AD achieves remarkable $F_1$ scores in the LunarLander dataset, with $99.8\%$ on policy anomalies and $98.0\%$ on perturbed anomalies. This shows our method is extremely effective in a noise-free and single-intention environment. We do not have results from GM-VASE in the LunarLander dataset because this method only works for the GPS dataset. 
\begin{table*}[t]
\begin{center}
\setlength\tabcolsep{5.5pt}
	\caption{Comparison of our method OIL-AD with baselines in the three datasets. The columns R, P and $F_1$ represent the recall, precision and $F_1$ score (as $\%$) respectively. For these three metrics, a higher value indicates a better performance.}
	\label{tab:main results}
	\begin{tabular}{l|ccc|ccc|ccc|ccc|ccc|ccc}\toprule
		\textit{Dataset} & \multicolumn{6}{c|}{\textit{LunarLander}} & \multicolumn{6}{c|}{\textit{Chengdu}} & \multicolumn{6}{c}{\textit{AIS}}  \\ 
  
    \textit{Anomaly} & \multicolumn{3}{c|}{\textit{Policy}} &\multicolumn{3}{c|}{\textit{Perturbed}}&\multicolumn{3}{c|}{\textit{Policy}} &\multicolumn{3}{c|}{\textit{Perturbed}}&\multicolumn{3}{c|}{\textit{Policy}} &\multicolumn{3}{c}{\textit{Perturbed}}\\
    \textit{Metric} & R &P &$F_1$ & R &P &$F_1$& R &P &$F_1$& R &P &$F_1$& R &P &$F_1$& R &P &$F_1$ \\ \midrule
    OC SVM
    &86.0 & 57.8&69.1&100.0 & 60.7&75.5
    & 89.2 & 47.7&62.2 &49.2 & 31.9&38.7&66.7 &10.9 &18.8 &58.7 & 9.88&16.9  \\

    LOF &15.2&41.8&22.2&100.0&85.7&92.3 &29.6&78.1&42.7&29.6&100.0&45.6&58.7&18.0&27.6&31.3&9.67&14.8\\

    IF & 32.4&15.7&21.1&100.0&38.0&55.0&31.2 &41.0&35.4&72.5&18.5&29.3&76.0&11.0&19.2&67.3&9.90&17.3\\

    LSTM-AE & 97.8&95.7&96.4&98.0&95.4&96.6 &75.2&54.1&62.8&25.8&24.2&24.8&84.0&19.3&31.3&59.3&14.3&23.1\\

    GM-VASE & -&-&-&-&-&-&79.6 &56.0&65.2&17.1&19.7&18.2&39.3&11.3&17.5&53.3&15.2&23.7\\
    ECOD & 7.20&7.02&7.10& 60.0&37.7&46.3&73.2 &15.3&25.3&95.2&18.5&31.1&48.0&18.2&26.4&25.3&10.6&15.0\\
    TS2vec &94.1&69.9&80.2&95.5&70.7&81.3&83.0&13.6&23.3& 97.0&16.3&28.0&66.7&9.38&16.5&73.3&10.2&18.0\\
    \hline
    OIL-AD &100.0 &99.6&\textbf{99.8}&99.1&96.8&\textbf{98.0}&84.0 &77.2&\textbf{80.4}&81.6&79.3&\textbf{80.4}&63.9&28.7&\textbf{39.6}&76.0&32.5&\textbf{45.5}\\

		  \bottomrule
	\end{tabular}
\end{center}
\end{table*}

\begin{figure*}[ht]
    \centering
    \subfloat[LunarLander\label{LunarLander}]{%
        \includegraphics[width=0.33\linewidth]{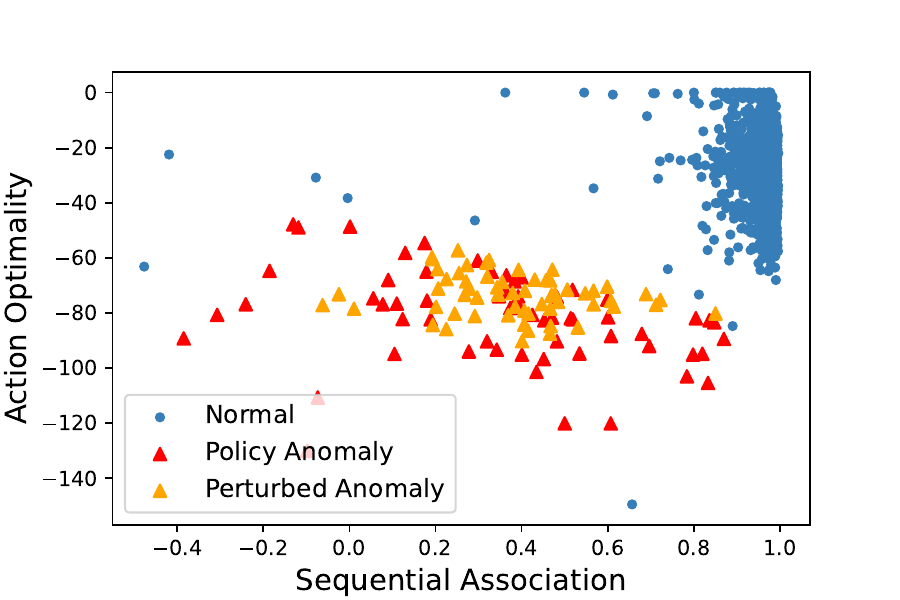}}
    \hfill
  \subfloat[Chengdu\label{Chengdu}]{%
       \includegraphics[width=0.33\linewidth]{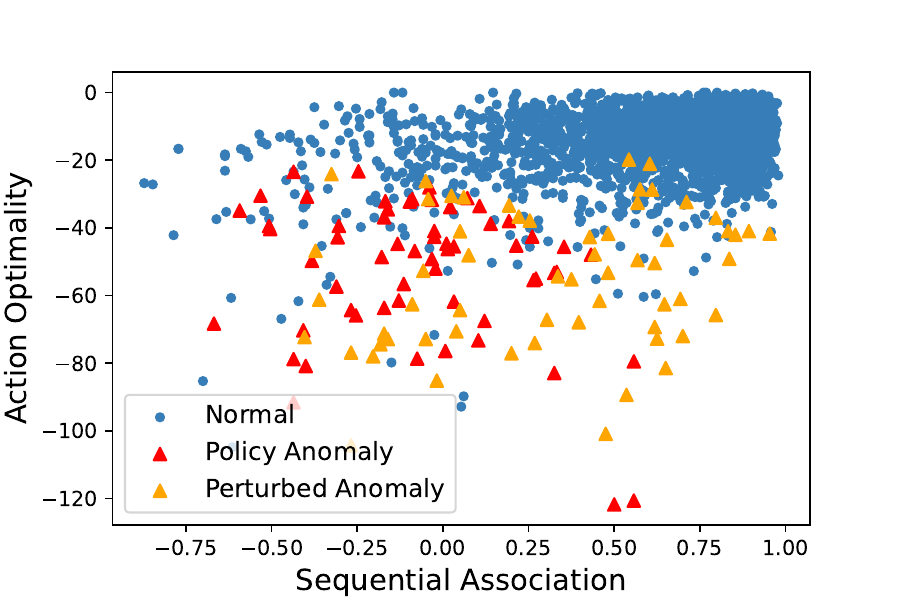}}
    \hfill
  \subfloat[AIS\label{AIS}]{%
        \includegraphics[width=0.33\linewidth]{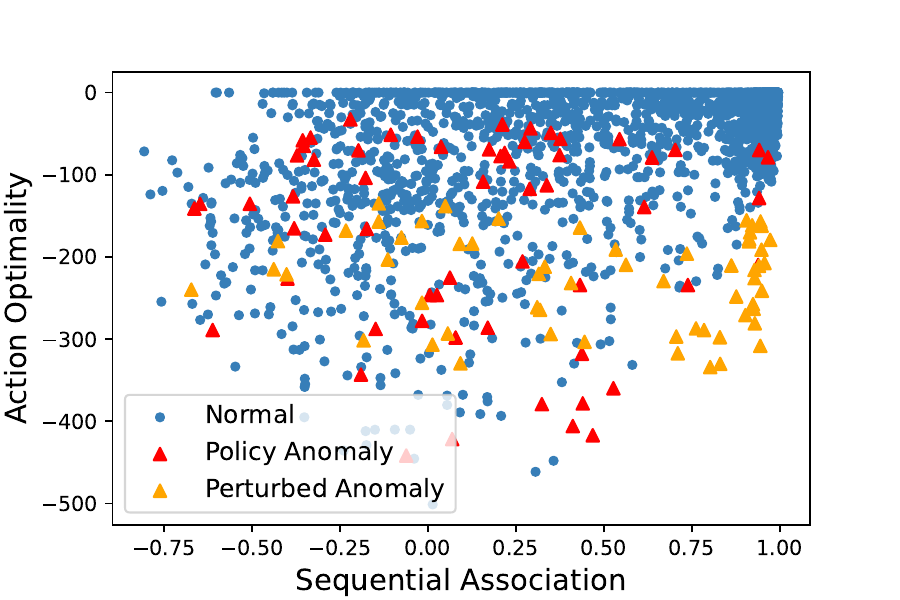}}
  \caption{2-D latent space of three datasets. 3000 normal features (blue) are randomly sampled from the training dataset. 60 policy anomalous features (red) and 60 perturbed anomalous features (orange) are randomly sampled from anomalous trajectories. The visualization is consistent with the results in Table \ref{tab:main results}.}
  \label{fig:visual}
\end{figure*}
\subsection{Evaluation on Latent Space}
We visualize the latent space of behaviour features on three different datasets as shown in Figure \ref{fig:visual}. We can see that behaviour features from normal trajectories cluster around the top right area indicating high action optimality and sequential association. In contrast, the anomalies have lower action optimality and sequential association. In particular, we can see that the normal features and the anomalous features are not well separated in the AIS dataset. This is due to the large variation of normal trajectories in terms of different destinations, different travel speeds and wide travel regions without the constraint of roads.

Comparing Table \ref{tab:main results} with Table \ref{tab:latent}, we can see that all the models achieve better performance in $F_1$ score by using the extracted behaviour features as inputs. Specifically, based on $F_1$ score, OC SVM achieves $37.6\%$ improvement ($38.7\%$$\to$$76.3\%$) on perturbed anomalies of the Chengdu dataset; LOF achieves $68.4\%$ improvement ($22.2\%$$\to$$90.6\%$) on policy anomalies of the LunarLander dataset; ECOD achieves $90.6\%$ improvement ($7.10\%$$\to$$97.7\%$) on policy anomalies of the LunarLander dataset.
This verifies the effectiveness of the latent space provided by our proposed behaviour features for anomaly detection. 
\begin{table*}[t]
\begin{center}
\setlength\tabcolsep{5.5pt}
	\caption{Comparison of different boundary models for OIL-AD in the three datasets. The inputs for all models are extracted behaviour features. The columns R, P and $F_1$ represent the recall, precision and $F_1$ score (as $\%$) respectively. For these three metrics, a higher value indicates a better performance.}
	\label{tab:latent}
	\begin{tabular}{l|ccc|ccc|ccc|ccc|ccc|ccc}\toprule
		\textit{Dataset} & \multicolumn{6}{c|}{\textit{LunarLander}} & \multicolumn{6}{c|}{\textit{Chengdu}} & \multicolumn{6}{c}{\textit{AIS}}  \\ 
  
    \textit{Anomaly} & \multicolumn{3}{c|}{\textit{Policy}} &\multicolumn{3}{c|}{\textit{Perturbed}}&\multicolumn{3}{c|}{\textit{Policy}} &\multicolumn{3}{c|}{\textit{Perturbed}}&\multicolumn{3}{c|}{\textit{Policy}} &\multicolumn{3}{c}{\textit{Perturbed}}\\
    \textit{Metric} & R &P &$F_1$ & R &P &$F_1$& R &P &$F_1$& R &P &$F_1$& R &P &$F_1$& R &P &$F_1$ \\ \midrule
    OC SVM&100.0&67.4&80.4&94.4&65.6&77.4  & 86.8&64.8&74.1&89.4&66.7&76.3&73.3&14.8&24.7&92.0&17.8&29.8\\

    LOF &99.6&83.0&90.6&93.6&80.7&86.7&69.8&63.4&66.4&74.8&60.8&67.0&78.0&12.2&21.1&67.3&10.7&18.4 \\
    ECOD &100.0&95.4&97.7&94.4 &95.2 &94.8&84.0&75.8&79.6&72.4&74.1&73.2&57.3 &18.5 &28.0&25.3&9.24&13.5\\
    IF (Ours)&100.0 &99.6&\textbf{99.8}&99.1&96.8&\textbf{98.0} &84.0 &77.2&\textbf{80.4}&81.6&79.3&\textbf{80.4}&63.9&28.7&\textbf{39.6}&76.0&32.5&\textbf{45.5}\\

		  \bottomrule
	\end{tabular}
\end{center}
\end{table*}
\subsection{Evaluation on Different Boundary Models}
Although we use Isolation Forest as the boundary model, there are other techniques to create a boundary on the latent space. In the following, we evaluate three alternatives to create boundaries in the two-dimensional latent space. In Table \ref{tab:latent}, we can see that Isolation Forest works the best to draw the boundary in terms of $F_1$ score. The reason might be that Isolation Forest can derive a more flexible boundary for normal features, resulting in a higher precision score.  
\subsection{Ablation Studies}
\begin{table}[t]
\begin{center}
	\caption{Ablation results ($F_1$ score) in training objectives. \textit{Obj 1} and \textit{Obj 2} refer to the action loss (Equation \ref{equ:first obj}) and the monotonicity loss (Equation \ref{equ:second obj}) respectively.}
	\label{tab:ablation}
	\begin{tabular}{llccc}\toprule
		Dataset &Anomaly& \textit{Obj 1} & \textit{Obj 2} & \textit{Obj 1+2} \\ \midrule
    \multirow{2}{*}{LunarLander}&Policy &87.0 &14.9&\textbf{99.8}\\
             &    Perturbed&66.2&27.7&\textbf{98.0}\\
             \hline
		\multirow{2}{*}{Chengdu} &Policy &66.2 &79.4&\textbf{80.4} \\
             &    Perturbed &53.6 &41.5&\textbf{80.4}           \\
            \hline
            \multirow{2}{*}{AIS} &Policy &30.6&\textbf{40.4}& 39.6   \\
            &    Perturbed &41.7&11.3 &\textbf{45.5}\\
            \hline
            
            Avg $F_1$ (as $\%$) & -&57.6 &35.9 &\textbf{74.0}\\
            
		 \bottomrule
	\end{tabular}
\end{center}
\end{table}
As shown in Table \ref{tab:ablation}, we further investigate the effect of each objective function in our model. We train the models purely based on Action Loss (\textit{Obj1}) or Monotonicity Loss (\textit{Obj2}) separately, and keep the rest of the detection setting the same.
The results show that our approach, by combining two objective functions, outperforms each individual objective function in terms of the average $F_1$ score. More specifically, using both objectives in the training process brings a $16.4\%$ increase in the average $F_1$ score. Note that taking only one of the objectives for training still achieves good performance compared with the other baselines in Table \ref{tab:main results}. In particular, \textit{Obj1} and \textit{Obj2} surpass the baselines in $F_1$ score by up to $18.0\%$ for perturbed anomalies on the AIS dataset and $14.2\%$ for policy anomalies on the Chengdu dataset respectively.
These verify that both objective functions are effective and necessary. 

\subsection{Parameter Sensitivity}
\begin{figure}[ht]
    \centering
    \subfloat{%
     \includegraphics[width=0.47\textwidth]{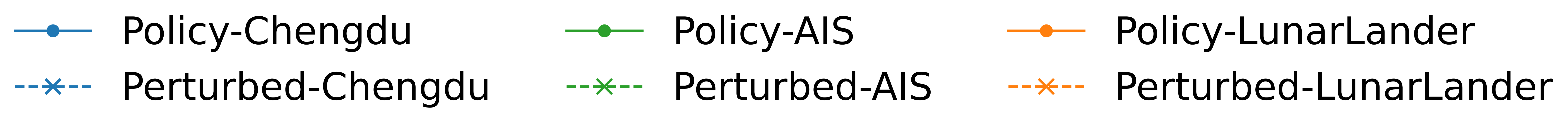}}
     \\
    \subfloat{%
       \includegraphics[width=0.235\textwidth]{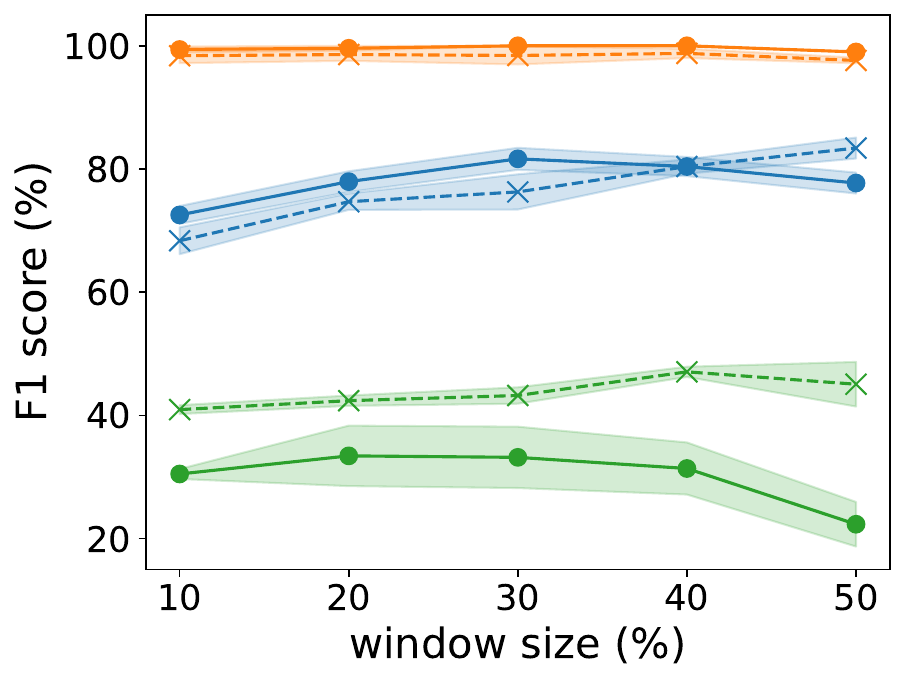} }
    \subfloat{%
    \includegraphics[width=0.235\textwidth]{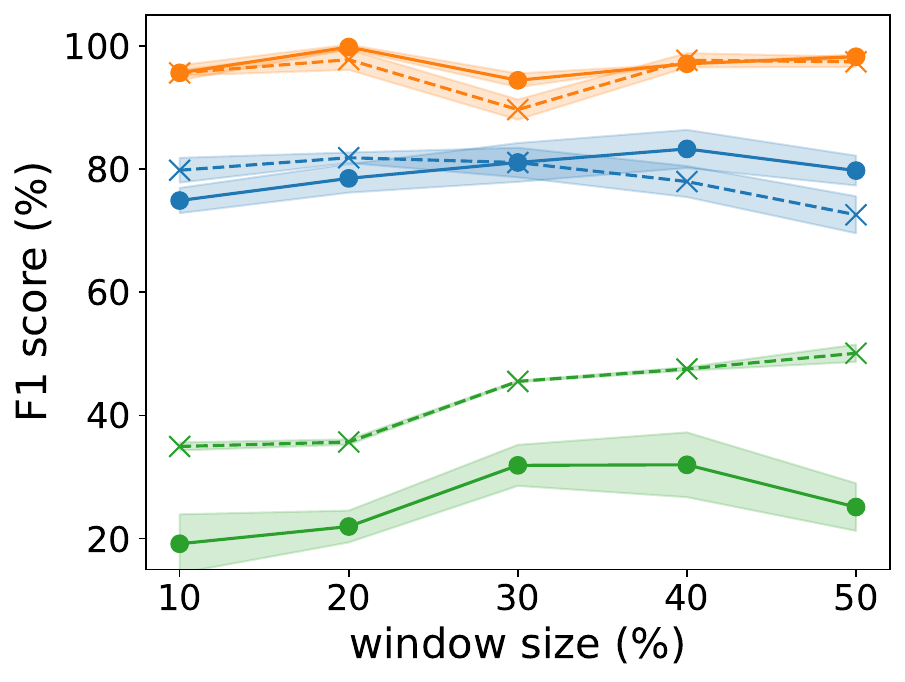} }
    \caption{Parameter sensitivity for sliding window sizes $w_q$ (left) and $w_v$ (right).}
    \label{fig:parameter}
\end{figure}

We provide the model performance under different choices for the sliding window sizes $w_q$ and $w_v$, as shown in Figure \ref{fig:parameter}. In particular, we set the sliding window sizes as different proportions (from $10\%$ to $50\%$) of the average length of normal trajectories for each dataset. Note that a smaller window size can be beneficial to detect the anomalies faster with less memory.
We can see that the relationship between the performance and the window sizes depends on the data pattern. For example, when the window size $w_v$ increases, the $F_1$ score of perturbed anomalies drops in the Chengdu dataset, but there is an increase in the AIS dataset. Generally, a larger window size of $w_q$ indicates a larger accumulation of the difference between normal decisions and anomalous decisions. This makes the performance drop for detecting policy anomalies while having the opposite effect on perturbed anomalies. The reason could be that the policy anomalous decisions are partial and concentrated in a short period, while perturbed anomalous decisions are spread throughout the trajectory. Meanwhile, a large window size $w_v$ tends to make small deviations imperceptible.

\subsection{Verification of the State Value Function}
The objective of the following experiment is to show that our algorithm is able to approximate the ground truth state value functions with only pre-recorded expert demonstrations, which verifies the effectiveness of the our monotonicity loss function (Eq.\ref{equ:second obj}). We evaluate the statistical correlations between recovered state values from our model and ground truth state values generated by First-visit Monte Carlo Prediction that utilises the reward signal to estimate the state values. By the law of large numbers, First-visit Monte Carlo can converge to $v_\pi(s)$ as the number of visits to each state goes to infinity \cite{sutton2018reinforcement}.
We compute two types of statistical correlations: Pearson's correlation coefficient (PCC) and Spearman's rank correlation coefficient (SCC). Higher PCC indicates higher linear correlations between two random variables and higher SCC suggests stronger monotonic relationships between two random variables. We choose the environment \textit{Taxi} from open AI Gym \cite{1606.01540}, which has discrete state and action spaces. The discrete environment guarantees the quality of the ground truth state values since it is very challenging to estimate the state value function in a continuous environment.  
The task in \textit{Taxi} includes navigating in a grid world, picking up the passengers and dropping them off at one of four designated destinations. 

Table \ref{tab: correlations} provides the PCC and SCC for anomalous agents that perform the trajectories with 6 randomly selected actions (around $31\%$ of the average trajectory length) and expert agents that follow the optimal policy. It can be seen that our algorithm can recover the state value function that has a high correlation with the ground truth. Even though we only have access to expert trajectories (state-action pairs) during training, our model also achieves high correlation in anomalous trajectories. Therefore, the recovered value function is sufficiently informative to represent the agents' goal.

\begin{table}
\begin{center}
\caption{Statistical correlations between recovered state values from OIL-AD and ground truth state values}
\label{tab: correlations}
\begin{tabular}{ c | c | c }\toprule
Metric & Anomalous Agents & Expert Agents\\\midrule

PCC& 0.745 & 0.945\\

SCC&0.708 &0.997\\ 
\bottomrule

\end{tabular}
\end{center}
\end{table}
\section{Conclusion}
In this work, we introduce Offline Imitation Learning based Anomaly Detection (OIL-AD) that uses a transformer based behavioural cloning policy network with a modified training process. The new training process combines both action loss and monotonicity loss to extract behaviour features for anomaly detection. The behaviour features, named action optimality and sequential association, construct an effective two-dimensional normality representation for detecting anomalous decision-making sequences. 
Unlike previous RL-based works, our learning process is purely offline, which means we do not need access to the reward function, environment dynamics or online interactions with the environment. This makes our method feasible for implementation in the real world. 

For future work, we will extend the current model to continuous action spaces. We will also consider different model structures, such as hierarchical neural networks, to incorporate sequential information. Apart from anomaly detection, the monotonicity loss of state values can also be instructive for RL-related tasks, such as planning \cite{silver2017predictron} and scheduling \cite{ren2021solving}.



\bibliographystyle{IEEEtranN}
\bibliography{main}

\newpage

\end{document}